\documentclass{article}
\usepackage{ijcai24}

\usepackage{times}
\usepackage{soul}
\usepackage{url}
\usepackage[hidelinks]{hyperref}
\usepackage[utf8]{inputenc}
\usepackage[small]{caption}
\usepackage{graphicx}
\usepackage{amsmath}
\usepackage{amsthm}
\usepackage{booktabs}

\usepackage[switch]{lineno}
\usepackage{graphicx}
\usepackage{float}
\usepackage{xcolor}
\usepackage{caption}
\usepackage{subcaption}
\usepackage{array}

\usepackage{algpseudocode}
\usepackage[ruled, linesnumbered]{algorithm2e}
\usepackage{dirtytalk}

\usepackage{upgreek}

\usepackage{acronym}
\usepackage[acronym]{glossaries}
\newacro{SSL}{Self-Supervised Learning} 
\newacro{CHRONOS}{Contrasting Heads Represent Opposed Natures of Signals}
\newacro{ML}{Machine Learning}
\newacro{ECG}{electrocardiogram} 
\newacro{AFib}{Atrial Fibrillation}
\newacro{DEBS}{Distilled Encoding Beyond Similarities}
\newacro{EEG}{electroencephalogram}
\newacro{CPC}{Contrastive Predictive Coding}
\newacro{PCLR}{Patient Contrastive Learning}
\newacro{EMA}{exponential moving average}
\newacro{SBnCL}{Subject-Based non Contrastive Learning}
\newacro{SVC}{ Support Vector Classificatier}
\newacro{BYOL}{Boostrap Your Own Latent}
\newacro{DINO}{Self-Distillation with no Labels}
\newacro{PAR}{Pondered Average Representation}
\newacro{SHHS}{Sleep Heart Health Study}
\newacro{ViT}{Vision Transformer}
\newacro{MLP}{Multilayer Perceptron}
\newacro{MIT-ARR}{MIT-BIH Arrhythmia Database}
\newacro{MIT-AFIB}{MIT-BIH Atrial Fibrillation Database}
\newacro{PCA}{Principal Component Analysis}
\newacro{MAE}{Masked Autoencoders}
\newacro{BYOL}{Bootstrap Your Own Latent}
\newacro{ReLU}{rectified linear unit}
\newacro{PSG}{Polysomnography}
\newacro{SOTA}{state-of-the-art}
\newacro{TF-C}{Time-Frequency Consistency}
\newacro{Cinc2017}{Physionet Challenge 2017}
\newacro{NSRR}{National Sleep Research Resource}
\newacro{CLOCS}{Contrastive Learning of Cardiac Signals Across Space}
\newacro{DEAPS}{Distilled Embedding for Almost-Periodic Time Series}
\newacro{SOTA}{state-of-the-art}
\newacro{VIC-REG}{Variance-Invariance-Covariance Regularization}
\newacro{SGD}{Stochastic Gradient Descent}
\newacro{LOO}{Leave-One-Out}

\title{Contrastive Learning Is Not Optimal for Quasiperiodic Time Series}

\author{
Adrian Atienza
\and
Jakob Bardram\and
Sadasivan Puthusserypady
\affiliations
Department of Health Technology,  Technical University of Denmark\\
\emails
\{adar, jakba, sapu\}@dtu.dk}
\begin{document}

\maketitle

\begin{abstract}
Despite recent advancements in \ac{SSL} for time series analysis, a noticeable gap persists between the anticipated achievements and actual performance.
% GAP
While these methods have demonstrated formidable generalization capabilities with minimal labels in various domains, their effectiveness in distinguishing between different classes based on a limited number of annotated records is notably lacking.
% Reason for the GAP
Our hypothesis attributes this bottleneck to the prevalent use of Contrastive Learning, a shared training objective in previous \ac{SOTA} methods. 
By mandating distinctiveness between representations for negative pairs drawn from separate records, this approach compels the model to encode unique record-based patterns but simultaneously neglects changes occurring across the entire record. 
% Intro to the method
To overcome this challenge, we introduce \ac{DEAPS} in this paper, offering a non-contrastive method tailored for quasiperiodic time series, such as \ac{ECG} data. 
% Why it solves the issue
By avoiding the use of negative pairs, we not only mitigate the model's blindness to temporal changes but also enable the integration of a \say{Gradual Loss ($\mathcal{L}_{gra}$)} function. This function guides the model to effectively capture dynamic patterns evolving throughout the record. 
The outcomes are promising, as \ac{DEAPS} demonstrates a notable improvement of +10\% over existing \ac{SOTA} methods when just a few annotated records are presented to fit a \ac{ML} model based on the learned representation.
    
\end{abstract}

\section{Introduction}
\begin{figure}[t]
    \centering
    \begin{subfigure}{0.485\linewidth}
        \centering
        \fbox{\includegraphics[width=\linewidth]{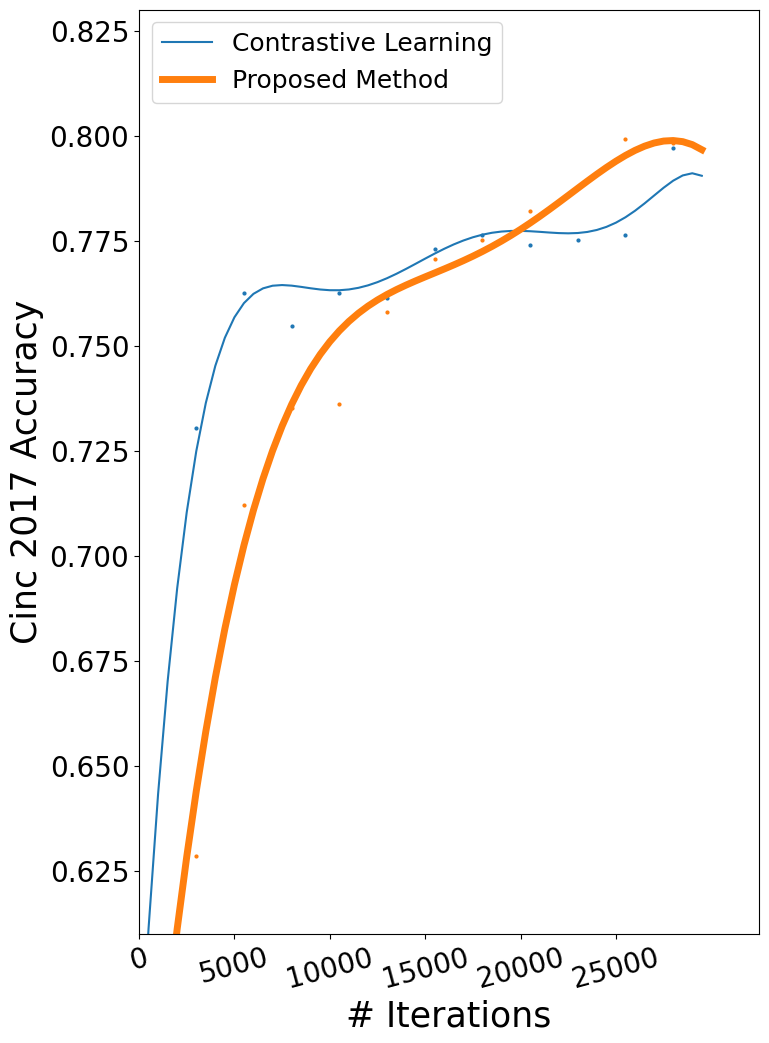}}
        \caption{Performance on Physionet Challenge 2017}
        \label{fig:cinc}
    \end{subfigure}
    \hfill
    \begin{subfigure}{0.47\linewidth}
        \centering
        \fbox{\includegraphics[width=\linewidth]{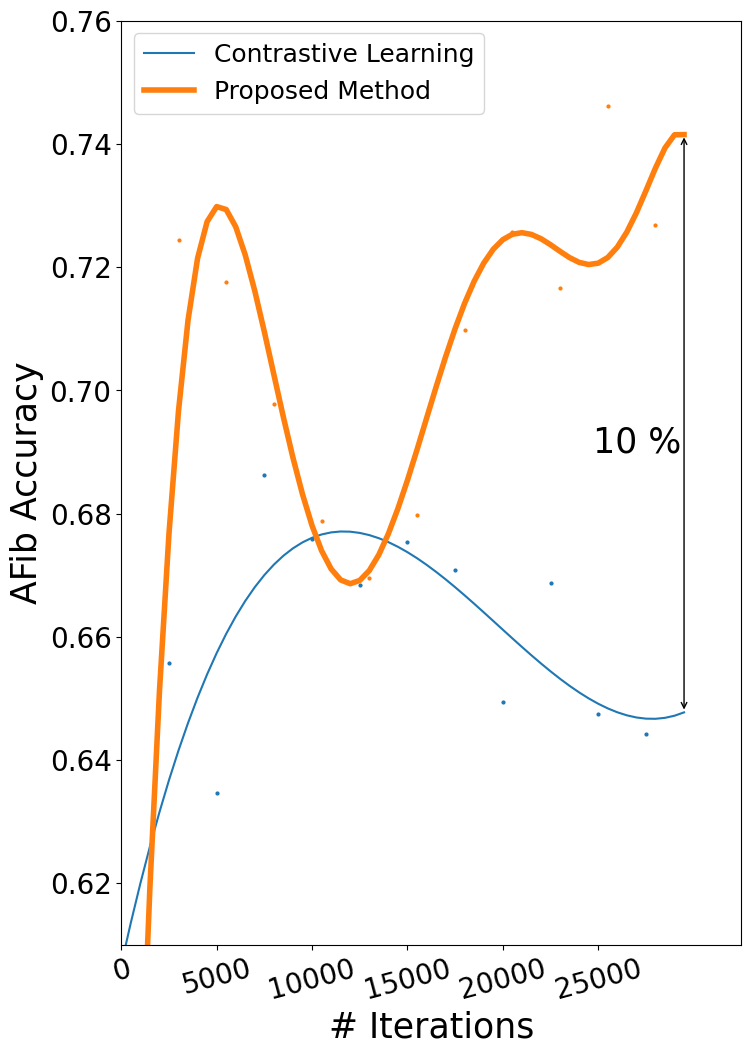}}
        \caption{From MIT-ARR Training to MIT-AFIB Testing}
        \label{fig:afib}
    \end{subfigure}
    \caption{Evolution of model performance across training procedures for the best performing Contrastive Learning Method (blue) and the proposed method (orange). While the downstream task displayed in Figure \ref{fig:cinc} considers multiple annotated recordings, the one shown in Figure \ref{fig:afib} only considers a few of them. For easier tracking of the evolution across iterations, a polynomial has been fitted.}
    \label{fig:task_evolution}
\end{figure}

With the rise of \acf{SSL}, there have been many efforts to adapt different works from other fields to time series processing, particularly physiological time series (\cite{pclr}, \cite{mix_up}). In this field, being able to optimize a model without the need for labels is highly valuable given that: (i) the model learns generic representations that are useful across many tasks, (ii) the labeling is costly, or the specific task is not known a priori, and (iii) the model learns representations that are more robust across different perturbations occurred during the recording of the data.
% What methods accomplish
Existing methods drive the model to learn representations that are refined throughout the optimization process. 
The learned representations are capable of performing a downstream task better and better throughout the training process, as shown in Figure \ref{fig:cinc},  when the \acf{ML} model considers a high number (8500) of recordings when is fitted. \\

%  What they dont (GAP)
\noindent In certain other fields of Artificial Intelligence, existing \ac{SSL} methods have shown promising results for a variety of applications by merely fitting a simple \ac{ML} model on top of the representation using a limited amount of data \cite{dino}, \cite{byol}. Achieving this level of performance is anticipated in a comparable time series analysis experiment, focusing on just a few records that encompass both normal and abnormal states. 
%It is to be expected to obtain this level of performance in a comparable experiment in Time Series processing, where just a few records containing both normal and abnormal states are considered. 
However, Figure \ref{fig:afib} shows that the results of this experiment are far from what was expected. It can also be seen how the performance of the model decreases throughout the optimization process.\\
    
\begin{figure}[H]
\centering
{\fbox{\includegraphics[width=0.95\linewidth]{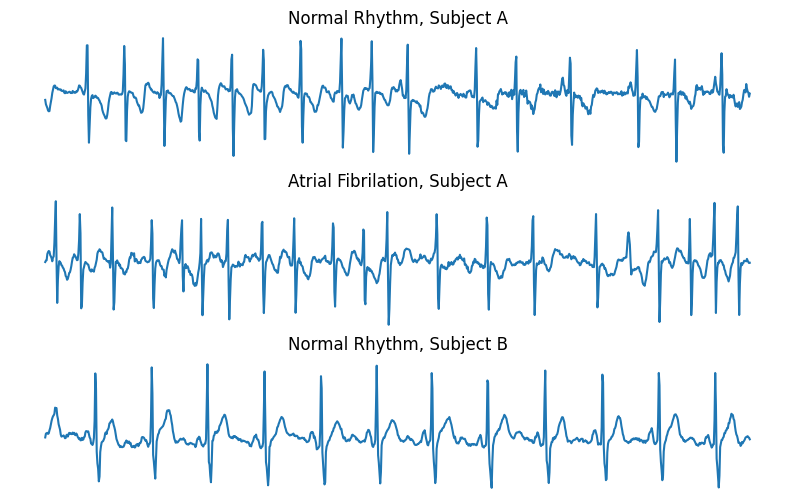}}}
\caption{A pair time series representing a normal and an abnormal state, belonging to the same subject are shown. In addition, another time strip that belongs to a distinct subject is displayed. The ECG morphology observed across various states within subject A exhibits consistent patterns, whereas this is different between subject A and B. These patterns are hypothesized to enable the model to categorize between two subjects while not reflecting subject state information.} 
\label{fig:ecgs}
\end{figure}

%Why they dont
\noindent While each method has its own uniqueness, they all share the use of Contrastive Learning \cite{simclr} as part of their optimization objective. This contrastive goal leads to the representations to be distant from the negative pairs which are drawn from distinct records. 
Quasiperiodic signals exhibit a combination of regular patterns with subtle variations. These quasiperiodic signals, such as \acf{ECG}, exhibit much more visible differences between recordings from different subjects than between different classes within the same recording (See Figure \ref{fig:ecgs}).\\

\noindent Just by seeking similarities between positive pairs belonging to the same record and dissimilarities between different ones, Contrastive learning drives the model to capture unique record-based representations.
However, in parallel, this contrastive objective leads the model to neglect the less visible, dynamic changes that occur across time.
This clarifies why these methods struggle to guide the model in generalizing across various classes present in a recording when provided with a limited set of recordings.

\noindent In this paper, we present \acf{DEAPS}, a new SSL method designed for physiological time series. The underlying concept of this approach is based on the categorization of signal patterns into two types: (i) Static patterns that account for individual characteristics such as gender and age and, (ii) dynamic patterns that can reveal transitional states or events experienced by the subjects during the recording. \ac{DEAPS} extends existing non-contrastive methods by not requiring the use of negative pairs during model optimization. This alleviates the model's blindness to dynamic patterns, while effectively continuing to encode the static ones.

\noindent In addition, avoiding the use of negative pairs allows \ac{DEAPS} to introduce a novel loss function, \say{Gradual Loss ($\mathcal{L}_{gra}$)}, that encourages the model to put more emphasis on representing the dynamic patterns. This loss function operates by following these steps: (i) Three time strips belonging to the same recording are given as inputs, (ii) Assuming that the dynamic patterns exhibit a smooth and linear behavior through the recording, $\mathcal{L}_{gra}$ ensures that the representation of the middle input can be interpolated from the other two representations, and (iii) A selective optimization is incorporated for only considering the most important features (i.e., the features for which their value changes the most over the recording interval).\\

\noindent We hypothesize that (i) the use of Contrastive Learning techniques, by introducing an objective that prioritizes that the representations of different recordings are far apart, causes the model to neglect the dynamic patterns, (ii) Dynamic patterns are much less evident than static ones, so avoiding the use of negative pairs is not enough. An additional objective function is needed for the model to capture them, and (iii) By effectively capturing these dynamic patterns, the model will acquire a better generalization capability by just considering a few labeled records.

\noindent As a result, \ac{DEAPS} is able to not only encode static patterns but also to distinguish different classes in the same recording by encoding the dynamic patterns too. Therefore, the model generalizes these classes to perform up to 10\% better when evaluated on another dataset, compared to SOTA contrastive methods (Figure \ref{fig:afib}). It also performs better when the ML model is fitted using multiple recordings (Figure \ref{fig:cinc}). In addition to these two experiments, \ac{DEAPS} is evaluated against SOTA methods in two other experiments. Overall, the evaluation involves up to 3 different tasks and 3 different datasets different from the one used during the optimization process.\\

\noindent In summary, the contributions of this paper are: 
\begin{enumerate}
    \item It is demonstrated, both logically and empirically, that Contrastive Learning objectives lead the model to neglect subtle changes within quasiperiodic time series data.
    
    \item We introduce \ac{DEAPS}, a novel \ac{SSL} method that stands apart from previous \ac{SOTA} Contrastive Learning methods. Dispensing with the negative pairs not only alleviates this model blindness to shifts within the record but also allows us to incorporate the novel $\mathcal{L}_{gra}$ function. This function, which emphasizes understanding changes within these quasiperiodic data, leads the model to encode dynamic patterns.
    
    \item We show that by understanding and encoding these dynamic patterns, the model is able to generalize different classes given a small sample of recordings where these two appear. This is exemplified by the up to 10\% improvement when compared with existing methods.
\end{enumerate}

\section{Related Work}
\begin{figure*}[!t]
\centering
{\fbox{\includegraphics[width=0.9\textwidth]{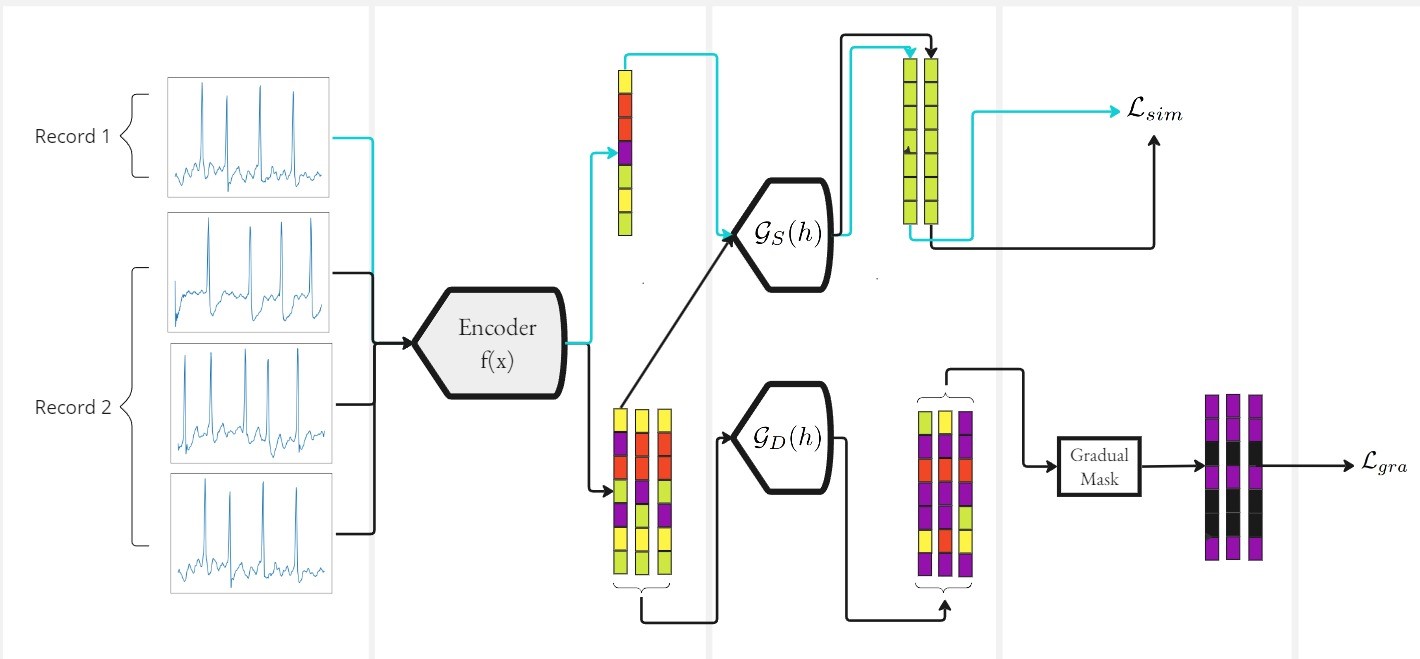}}}
\caption{DEAPS illustrated. Different colors represent different temporal patterns: Green (static patterns), and purple (dynamics patterns). The different colors of the arrows indicate inputs from different recordings. While $\mathcal{L}_{sim}$ is computed between the time series representations belonging to different records, $\mathcal{L}_{gra}$ is computed using the time series belonging to the same record. The four inputs belong to the same subject. The encoder is the only component that is saved at the end of the training procedure. The other components are dismissed.} 
\label{fig:deaps}
\end{figure*}

\subsubsection{Unsupervised Learning in Time Series}
Interpreting physiological signals demands a level of expertise, making the labeling of such data costly. It represents a bottleneck within the deep learning context, particularly given the substantial volume of data required for training deep learning models. Consequently, various studies have been directed towards optimizing the model using a limited amount of labels \cite{unsupervised2} or directly without relying on these labels (\cite{unsupervised1}, \cite{unsupervised3}).

\subsubsection{Contrastive Learning in Time Series}
With the rise of \ac{SSL}, different methods have recently been designed in the field of time series analysis, exploiting the characteristics of the time series data. 
\ac{CLOCS} \cite{clocs} utilizes two consecutive \ac{ECG} time strips as positive pairs. 
The mixing-up method~\cite{mix_up} utilizes the temporal characteristics of the data for a more tailored data augmentation by creating a time series, which is a product of two time series from the same subject. 
An alternative approach is proposed in the \ac{TF-C} \cite{tfc} method, which produces two variations of the time-domain and frequency-domain pairs associated with the input signal. \ac{SSL} method is developed to identify similarities inherent in these paired representations.
The last example is \ac{PCLR} \cite{pclr} which considers two inputs from the same subject but different recordings as positive pairs.

Although each of these methods has its particularity in defining the positive pairs, they all share the use of the Contrastive Learning objective. 
\ac{PCLR} obtains the best results in the benchmark used in this work. Accordingly, \ac{DEAPS} considers the same selection of positive pairs criteria, i.e., between pairs from different recordings.

\subsubsection{Non-Contrastive Learning}
In other fields of artificial intelligence, such as Computer Vision, non-Contrastive Methods have emerged as an alternative to Contrastive Methods. The major challenge of these methods is to avoid the collapse mode in the representations. This collapse occurs when the model computes the same representations independently of the given input. Given that non-contrastive methods do not force representations of negative pairs to be distinct, other strategies have to be considered to avoid this issue.

Methods such as \ac{BYOL} \cite{byol} or \ac{DINO} \cite{dino} rely on a teacher-student architecture. The cost functions are computed between the corresponding teacher-student. While the student network is optimized with \ac{SGD}, the weights of the teacher network are updated via \ac{EMA} of the student weights. Other methods such as \ac{VIC-REG} \cite{vicreg} maximize the batch variance of each feature within the representations. 
The method presented in this work utilizes \ac{BYOL} framework as the baseline.

\section{\acf{DEAPS}}
\ac{DEAPS} is designed to capture both the static and the dynamic patterns of the time series. It is illustrated in Figure~\ref{fig:deaps}.
The model takes up to four different time strips belonging to two different records from the same subject. 
The encoder computes the representations of these distinct inputs.
The static projector (labeled $\mathcal{G}_S$) takes a pair of representations from the two different recordings which are displayed in a static space.
The similarity loss function ($\mathcal{L}_{sim}$) is computed between these pairs of projections, ensuring that the unique patterns of each subject are represented consistently even between different records. 
The dynamic projector (labeled $\mathcal{G}_D$) processes the three representations from the same recording.
\ac{DEAPS} only considers the most important features for each triplet of inputs, i.e., the ones that are encoding dynamic patterns that are evolving between this time interval. The rest of the dynamic projection is masked.
The unmasked projections are used for computing the gradual loss function ($\mathcal{L}_{gra}$).
The overall loss function proposed by \ac{DEAPS} can be expressed as the following:
\begin{equation}
    \mathcal{L}_{\textit{DEAPS}} = \mathcal{L}_{sim} + \mathcal{L}_{gra} + \alpha \cdot  \mathcal{L}_{c}.
\end{equation}
This section provides detailed explanations for each of its different components.

\subsection{Non-Contrastive Learning. Revisiting BYOL}
\ac{BYOL} \cite{byol} was the first \ac{SSL} non-contrastive learning method that dispensed with negative pairs. \ac{DEAPS} utilizes it as baseline. 

In other to avoid the mode collapse when the use of negative pairs is removed, \ac{BYOL} features both a teacher network and a student network, each with an encoder and a projector initialized with identical parameters. The student network is additionally equipped with a predictor, which acts on the views computed by the projector. The optimization involves \ac{SGD} for the student network, projector, and predictor. Both the teacher network and the teacher projector serve as an \ac{EMA} of the student network. The loss function only measures the similarities between the student/teacher output pairs and it is described as the following:

\begin{equation}
\resizebox{.81\linewidth}{!}{$
            \displaystyle
            \mathcal{L}_{sim}(\mathbf{z}_2^s, \mathbf{q}(\mathbf{\upzeta}_1)^s) = 1 - \frac{\mathbf{z}_2^s \cdot \mathbf{q}(\mathbf{\upzeta}_1)^s}{\max \left(\left\|\mathbf{z}_2^s \right\|_2 \cdot\left\|\mathbf{q}(\mathbf{\upzeta}_1)^s\right\|_2, \epsilon\right)}, 
$}
\end{equation}
where $\mathbf{z}_2^s$ and $\mathbf{q}(\mathbf{\upzeta}_1)^s$ are the teacher similarity projection and student similarity prediction, respectively, for inputs belonging to the first and second records.

Both the projector and the predictor are discarded after the optimization process. Therefore, the representations used for the downstream tasks are obtained directly from the student network. The student-teacher framework and the characteristic predictor are excluded from Figure \ref{fig:deaps} for clarity. In practice, the different loss functions are computed between projection/prediction pairs obtained by the teacher and student networks, denoted as $\mathbf{z}$ and $ \mathbf{q}(\mathbf{\upzeta})$, respectively, where $\upzeta$ represents the student projection.

\subsection{Gradual Loss} \label{sec:gra_loss}
Avoiding the use of a Contrastive objective alleviates the model's blindness to changes in the record. However, it is not sufficient to lead the model to capture dynamic patterns. Non-contrastive methods do not compare each representation with the rest of the representations in the batch. It allows us to incorporate a metric that compares just the representations within the same record.
% It is crucial to emphasize that all four inputs used for the optimization originate from the same subject and hence are all positively linked. This is the reason that we consider them as positive pairs.

\subsubsection{Capturing the Dynamic Patterns} 
We introduce the \say{Gradual Loss ($\mathcal{L}_{gra}$)} as a part of the training objective. This loss function just considers $X_{t - i}$, $X_{t}$, and $X_{t+j}$, three that belong to the same record and have been drawn over a fixed window size.
DEAPS enforces the representation of the dynamic patterns to exhibit a linear behavior by mandating an accurate interpolation of the representation of an inter-
mediate input from the representations of the other two. In other words, if a subject’s state evolves from $X_{t - i}$ to $X_{t+j}$, the representation of $X_t$, i.e., $z_t$, should approximate an intermediate point between $z_{t - i}$ to $z_{t+j}$. This objective ensures that the temporal evolution is properly captured within the representations. $\mathcal{L}_{gra}$ is described as:

\begin{equation}
    \resizebox{.91\linewidth}{!}{$
            \displaystyle
            \mathcal{L}_{gra}(\mathbf{q}_s(\mathbf{z}_{t}^s), \mathbf{z}_{t - i}^t, \mathbf{z}_{t + j}^t) = 1 - \frac{\mathbf{q}_s(\mathbf{z}_{t}^s) \cdot \mathcal{PAR}(\mathbf{z}_{t - i}^t, \mathbf{z}_{t + j}^t)}{\max \left(\left\|\mathbf{q}_s(\mathbf{z}_{t}^s)\right\|_2 \cdot\left\|\mathcal{PAR}(\mathbf{z}_{t - i}^t, \mathbf{z}_{t + j}^t)\right\|_2, \epsilon\right)},
        $}
\end{equation}
where \ac{PAR} is the approximation of $\mathbf{z}_{t}$, drawn from $\mathbf{z}_{t-i}$ and $\mathbf{z}_{t+j}$. We do not force $\mathbf{z}_{t}$ to be equally distant from both $\mathbf{z}_{t-i}$ and $\mathbf{z}_{t+j}$, therefore, we calculate \ac{PAR} as,

\begin{equation}
\resizebox{.53\linewidth}{!}{$
\mathcal{PAR}(\mathbf{z}_{t - i}^t, \mathbf{z}_{t + j}^t)=\frac{\mathbf{z}_{t - i}^t \cdot j + \mathbf{z}_{t + j}^t \cdot i} {i + j}.
$}
\end{equation}

Note that the incorporation of $\mathcal{L}_{gra}$ into the method is only possible since \ac{DEAPS} is a non-contrastive method. $\mathcal{L}_{gra}$ only makes sense when taking into account only the $\mathbf{z}_{t}$ and the interpolation of this representation given the $\mathbf{z}_{t - i}$ and $\mathbf{z}_{t + j}$ pairs, and not the rest of the representations in the batch considered as negative-pairs.

\subsubsection{Intuitions of the Use of Two Projectors}
$\mathcal{L}_{sim}$ and $\mathcal{L}_{gra}$ have conflicting goals. While the first one prioritizes that representations between positive pairs are invariant, the purpose of $\mathcal{L}_{gra}$ is encouraging changes within the recording to be captured in the representations. To facilitate the disentanglement between static and dynamic features, two projectors (represented as $\mathcal{G}_{s}$ and $\mathcal{G}_{d}$ in Figure \ref{fig:deaps}) are incorporated into both the student and teacher network instead of the single projector of the original \ac{BYOL} framework. In this way, only static features will be projected by $\mathcal{G}_{s}$ and thus taken into account by $\mathcal{L}_{sim}$. Analogously, only dynamic features will be considered when computing $\mathcal{L}_{gra}$. 

\subsubsection{Intuitions of the Windows Size}
An essential consideration in implementing the method is determining the appropriate spacing window size between $\mathbf{X}_{t-i}$ and $\mathbf{X}_{t+j}$, i.e., how much these inputs may be separated in time. This spacing window size must be large enough to accommodate signal changes, yet narrow enough just allow different patterns that encode distinct changes to move in one single direction each. If successive changes occur within this window, it can lead to conflicting directions in which these changes are reflected in the representations, thereby $\mathcal{PAR}(\mathbf{z}_{t - i}^t, \mathbf{z}_{t + j}^t)$ may not be aligned with $\mathbf{q}_s(\mathbf{z}_{t}^s)$. The optimal value of this window size may vary depending on the quasiperiodic data modality. The effect of this size is studied in Section~\ref{sec:ablation}.

\subsection{Selective Optimization} \label{sec_sel_opt}
It is assumed that a set of features contained in the representation encodes the dynamic patterns of the time series. However, it is not reasonable to think that all these patterns will evolve in every fixed time window.
To address this issue, \ac{DEAPS} introduces a Selective Optimization. This ensures that only the features that encode changing patterns within a particular time window are considered when computing the loss function.

\paragraph{Masking the Output} The process of determining which features of each dynamic projection to include in each loss function follows these steps: (i) It calculates the absolute difference between the values dynamic projections of each input pair belonging to the $\mathbf{X}_{t-i}$ and $\mathbf{X}_{t+j}$ batches, (ii) the \textit{N} values that exhibit a larger difference are considered when computing $\mathcal{L}_{gra}$, and (iii) any remaining features in each projection are masked and their gradients are not considered. This process is represented in Figure \ref{fig:deaps}. The effect of this selective optimization is discussed in section~\ref{sec:ablation}.

\paragraph{Covariance Term} To enhance the efficacy of our selective optimization approach, we strive to ensure that each feature within the representations encapsulates a distinctive and individual pattern. To achieve this objective, we incorporate the Covariance Loss function ($\mathcal{L}_c$) as a regularization factor, effectively penalizing redundancy within the representations. This particular cost function, which has also found application in prior studies like Barlow Twins \cite{btwins} and \ac{VIC-REG} \cite{vicreg}, operates as follows:

\begin{equation}
\mathcal{L}_c(\upzeta)=\frac{1}{d} \sum_{i \neq j}[C(\mathbf{\upzeta})]_{i, j}^2,
\end{equation}
where $C(\mathbf{\upzeta})$ is the covariance matrix computed on the student projection, $\mathbf{\upzeta}$. Its effect is studied in Section \ref{sec:ablation}.

\subsection{Implementation Details}
In order to ensure the reproducibility of the method, this subsection details the configuration of the hyperparameters. 

\paragraph{Model Architecture} We use an adaptation of the \ac{ViT}~\cite{vit} model for processing physiological signals. The input data is a time series of 1000 samples, which correspond to 10 seconds-length signal sampled at 100Hz. This input is split into segments of a length of 20 samples. The model counts with 6 regular transformer blocks with 4 heads each. The model dimension is set to 128, for a total of 1,192,616 trainable parameters.

\paragraph{\ac{DEAPS} Implementation} The projectors and predictors in our approach are implemented as a two-layer \ac{MLP} with a dimensionality of 512 and 256, respectively. Batch normalization and \ac{ReLU} operations are incorporated between the two layers of each structure. The \ac{EMA} updating factor ($\tau$) is set to 0.995. The window size is set to 2 minutes. We weigh the covariance loss with a factor of 0.1. We optimize the most important 32 features during the selective optimization. The effect of both the window size and the number of features to be considered during the selective optimization is discussed in Section~\ref{sec:ablation}.

\paragraph{Optimization} The model is trained with 10 second-length signals belonging to the \ac{SHHS} dataset~\cite{shhs1}, \cite{shhs2}. The training procedure consists of 30,000 iterations. We use a batch size of 256, and Adam \cite{adam} with a learning rate of $3e-4$  and a weight decay of $1.5e-6$ as the optimizer. The training procedure and the evaluations are performed on a local computer, with a Nvidia GeForce RTX 3070 GPU.

\section{Evaluation}\label{sec:eval}
The proposed model has undergone extensive simulation studies to evaluate its performance.
%To confirm the validity of our model, we conducted an extensive evaluation. 
Initially \ac{DEAPS} was assessed by comparing it to four \ac{SOTA} methods across three distinct downstream tasks, each using four different databases. These four databases are: \ac{MIT-AFIB}\cite{mit-afib}, \ac{MIT-ARR}, \cite{mit-afib}, \ac{Cinc2017} \cite{cinc2017} and \ac{SHHS} \cite{shhs1}. All used databases are publicly available in Physionet \cite{physionet} and \ac{NSRR}.\\

\noindent Furthermore, we conducted a \ac{PCA} analysis to substantiate the core concept of our approach, which revolves around disentangling static and dynamic patterns within time-series data with the aim of demonstrating DEAPS' capability to encode both of them.

\subsection{Comparison against \ac{SOTA} Methods}

In order to evaluate the performance of DEAPS, three experiments of different nature have been carried out: (i) \ac{AFib} identification to assess the effectiveness of generalising different classes within the same recording (ii) Gender Identification, to assess the effectiveness of static patterns and (iii) Physionet Challenge 2017 to assess the scalability of the representations when multiple recordings are given.
\ac{DEAPS}' performance has been compared against the four most relevant \ac{SOTA} methods, namely; (i) \ac{PCLR}~\cite{pclr}, (ii) \ac{CLOCS} \cite{clocs}, (iii) Mixing-Up \cite{mix_up}, and (iv) \ac{TF-C} \cite{tfc}.  
Moreover, we have included the standard \ac{BYOL}~\cite{byol} framework as a baseline.\\

\noindent To ensure fairness in evaluation, we have optimized the same model used in this work, under the same configuration (optimizer, data, batch size, and number of iterations), except for the TF-C method, where their proposed model has been used, since it requires the use of two encoders instead of one. The TF-C model contains approximately 32 million parameters, which is 30x more parameters than the proposed \ac{DEAPS} model. Therefore, it has been optimized over 75K iterations, instead of the 30K iterations proposed in this work.

\subsubsection{\acf{AFib} Identification} 
In order to assess the ability of the method to generalise different classes within the same record, we have conducted two experiments where there is no subject overlap between the training and validation sets. This overlapping would significantly simplify AFib identification. In the first experiment, we employed 10-second \ac{ECG} time series strips originating from the eight subjects affected by AFib, belonging to the \ac{MIT-ARR} database for fitting a \ac{SVC} \cite{svc}  on top of the representation. We subsequently evaluated this model on the complete \ac{MIT-AFIB} database. In the second, we have conducted a \ac{LOO} validation across the 23 \ac{MIT-AFIB} subjects. The outcomes of this experiment are tabulated in Table \ref{tab:afib}. \ac{DEAPS} demonstrates a significantly superior performance.

\begin{table}[H]
    \resizebox{\columnwidth}{!}{%
        \begin{tabular}{c|ccc|c|}
            \cline{2-5}
            \multicolumn{1}{l|}{} & \multicolumn{3}{c|}{\textbf{MIT-ARR $\rightarrow$ MIT-AFIB}} & \textbf{\begin{tabular}[c]{@{}c@{}}MIT-AFIB\\ LOO\end{tabular}} \\ \hline
            \multicolumn{1}{|c|}{\textbf{\begin{tabular}[c]{@{}c@{}}SSL \\ METHOD\end{tabular}}} & \textbf{ACCURACY (\%)} & \textbf{SENSITIVITY (\%)} & \textbf{SPECIFICITY (\%)} & \textbf{ACCURACY (\%)} \\ \hline
            \multicolumn{1}{|c|}{Mixing-Up} & 65.6 & 60.6 & 67.4 & 73.4$\pm$17.1 \\
            \multicolumn{1}{|c|}{TF-C} & 71.8 & 64.8 & 76.5 & 77.2$\pm$20.5 \\
            \multicolumn{1}{|c|}{PCLR} & 65.5 & 59.8 & 68.1 & 75.1$\pm$18.2 \\
            \multicolumn{1}{|c|}{BYOL} & 67.9 & 62.6 & 69.9 & 75.6$\pm$21.1 \\
            \multicolumn{1}{|c|}{CLOCS} & 62.8 & 55.0 & 66.2 & 75.4$\pm$19.1 \\
            \multicolumn{1}{|c|}{\textbf{DEAPS}} & \textbf{75.5} & \textbf{74.6} & \textbf{75.9} & \textbf{79.4$\pm$18.7} \\ \hline
        \end{tabular}%
    }
\caption{AFib Identification given a few, non-overlapping annotated records. In MIT-ARR $\rightarrow$ MIT-AFIB experiment. a SVC is fitted and evaluated on these datasets, respectively. In MIT-AFIB LOO experiment, a Leave-One-Out cross-validation is carried out.}

\label{tab:afib}
\end{table}

\subsubsection{Gender Classification} 

We randomly selected 1549 \ac{ECG} time series strips of length 10 seconds, each associated with a distinct subject, from the \ac{SHHS} database. We conducted a five-fold cross-validation to evaluate the performance of the downstream tasks. The \ac{ML} model used is the \ac{SVC} and the outcomes of these experiments are presented in Table \ref{tab:gender_table}, where it can be seen how \ac{DEAPS} obtains comparable results with the best performing method. 
\begin{table}[H]
    \centering
    \resizebox{0.8\columnwidth}{!}{%
        \begin{tabular}{c|ccc|}
            \cline{2-4}
            \multicolumn{1}{l|}{} & \multicolumn{3}{c|}{\textbf{GENDER IDENTIFICATION}} \\
            \hline
            \multicolumn{1}{|c|}{\textbf{\begin{tabular}[c]{@{}c@{}}SSL \\ METHOD\end{tabular}}} & \textbf{\begin{tabular}[c]{@{}c@{}}TOTAL \\ ACCURACY (\%)\end{tabular}} & \textbf{\begin{tabular}[c]{@{}c@{}}MALE \\ ACCURACY (\%)\end{tabular}} & \textbf{\begin{tabular}[c]{@{}c@{}}FEMALE \\ ACCURACY (\%)\end{tabular}} \\
            \hline
            \multicolumn{1}{|c|}{Mixing-Up} & 70.4$\pm$1.5 & 68.0$\pm$3.3 & 71.9$\pm$2.3 \\
            \multicolumn{1}{|c|}{TF-C} & 65.8$\pm$2.9 & 62.8$\pm$5.5 & 67.4$\pm$2.1 \\
            \multicolumn{1}{|c|}{PCLR} & 76.1$\pm$1.7 & 74.7$\pm$2.6 & 77.2$\pm$3.1 \\
            \multicolumn{1}{|c|}{BYOL} & \textbf{76.3$\pm$3.3} & \textbf{75.3$\pm$4.0} & \textbf{77.6$\pm$3.9} \\
            \multicolumn{1}{|c|}{CLOCS} & 70.4$\pm$1.11 & 68.1$\pm$3.1 & 71.9$\pm$2.2 \\
            \multicolumn{1}{|c|}{\textbf{DEAPS}} & 75.8$\pm$1.1 & 74.4$\pm$2.9 & 76.6$\pm$2.5 \\
            \hline
        \end{tabular}%
    }
    \caption{Evaluation of gender classification task. Five-cross evaluation for the SVC model fitted on top of the representations.}
    \label{tab:gender_table}
\end{table}

\subsubsection{Physionet Challenge 2017} In this final experiment, we utilized the Cinc2017 database, which categorizes \ac{ECG} time series strips as either Sinus Rhythm (SR), AFib, or Others. We used the dataset's pre-defined partitioning of ``train'' and ``validation'' sets for evaluating the \ac{SVC} model fitted on top of the representations. The findings of this experiment are summarized in Table \ref{tab:physio}, wherein DEAPS achieves superior performance compared with the rest of the methods.
\begin{table}[H]
\resizebox{\columnwidth}{!}{%
\begin{tabular}{c|cccc|}
\cline{2-5}
\multicolumn{1}{l|}{}                                                                & \multicolumn{4}{c|}{\textbf{\begin{tabular}[c]{@{}c@{}}PHYSIONET  CHALLENGE\\  2017\end{tabular}}}                                                                                                                                                                                                                 \\ \hline
\multicolumn{1}{|c|}{\textbf{\begin{tabular}[c]{@{}c@{}}SSL \\ METHOD\end{tabular}}} & \multicolumn{1}{c|}{\textbf{\begin{tabular}[c]{@{}c@{}}TOTAL\\ ACCURACY (\%)\end{tabular}}} & \textbf{\begin{tabular}[c]{@{}c@{}}SR\\ ACCURACY (\%)\end{tabular}} & \textbf{\begin{tabular}[c]{@{}c@{}}AFIB\\ ACCURACY (\%)\end{tabular}} & \textbf{\begin{tabular}[c]{@{}c@{}}OTHER\\ ACCURACY (\%)\end{tabular}} \\ \hline
\multicolumn{1}{|c|}{Mixing-Up}                                                      & \multicolumn{1}{c|}{74.9}                                                                   & 74.3                                                                & 85.4                                                                  & 71.0                                                                   \\
\multicolumn{1}{|c|}{TF-C}                                                           & \multicolumn{1}{c|}{60.2}                                                                   & 62.0                                                                & \textbf{100}                                                          & 44.3                                                                   \\
\multicolumn{1}{|c|}{PCLR}                                                           & \multicolumn{1}{c|}{77.5}                                                                   & 75.0                                                                & 88.3                                                                  & 79.5                                                                   \\
\multicolumn{1}{|c|}{BYOL}                                                           & \multicolumn{1}{c|}{78.0}                                                                   & 76.5                                                                & 84.5                                                                  & 79.0                                                                   \\
\multicolumn{1}{|c|}{CLOCS}                                                          & \multicolumn{1}{c|}{74.8}                                                                   & 72.7                                                                & 84.7                                                                  & 77.1                                                                   \\
\multicolumn{1}{|c|}{\textbf{DEAPS}}                                                 & \multicolumn{1}{c|}{\textbf{80.1}}                                                          & \textbf{77.4}                                                       & 86.0                                                                  & \textbf{85.2}                                                          \\ \hline
\end{tabular}}
\caption{Evaluation on Physionet Challenge 2017. The original train-validation split has been used when fitting the SVC model.}
\label{tab:physio}
\end{table}

\subsection{PCA Analysis on Representations} The fundamental concept underpinning this approach revolves around the integration of dissimilarities to facilitate the encoding of both static and dynamic features within the model. To ascertain the validity of this premise, a Principal Component Analysis (PCA) \cite{pca} is executed on the representations from \ac{MIT-AFIB} database.\\ 

\begin{figure}[t]
\centering

\subfloat[Subject-Based Analysis]
{\fbox{\includegraphics[width=0.48\linewidth]{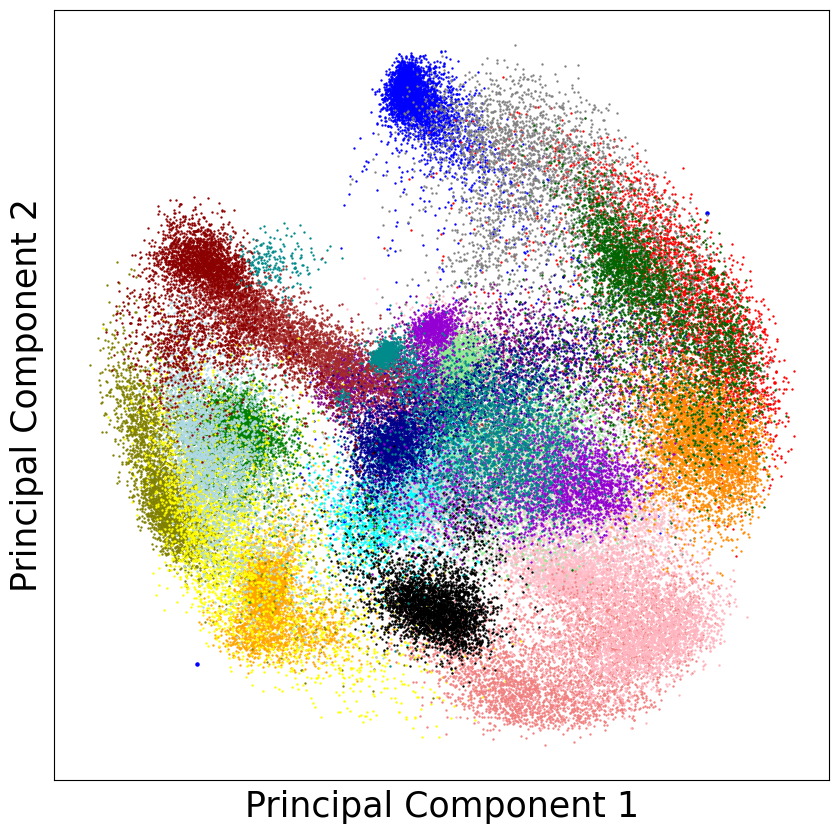}\label{fig:static_pca}}}
\hfill
\subfloat[Episode-Based Analysis]
{\fbox{\includegraphics[width=0.43\linewidth]{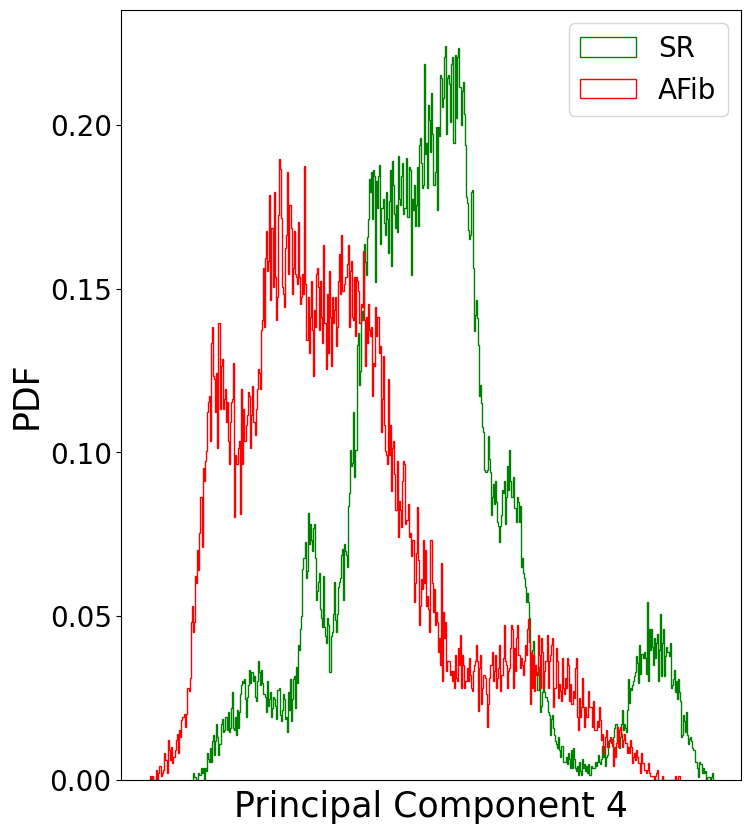}\label{fig:pca_afib}}}
\caption{A PCA has been fitted on top of the MIT-AFIB representations. The values of the different components have been studied for disentangling the static and dynamic features.}
\label{fig:pca}
\end{figure}

\noindent As depicted in Figure \ref{fig:static_pca} where different patients are displayed in different colors, temporal segments originating from the same individual exhibit analogous positions in Principal Components 1 and 2, irrespective of the individual's status. Conversely, Principal Component 5 yields distinct values between AFib and normal rhythm (NR) classifications, regardless of the individual, as shown in Figure \ref{fig:pca_afib}. This second behavior is not seen in Contrastive Learning methods (See Appendix).

\begin{figure*}[t]
\centering
\resizebox{\textwidth}{!}{%
\subfloat[MIT-ARR $\rightarrow$ MIT-AFIB]
{\fbox{\includegraphics[width=0.5\textwidth]{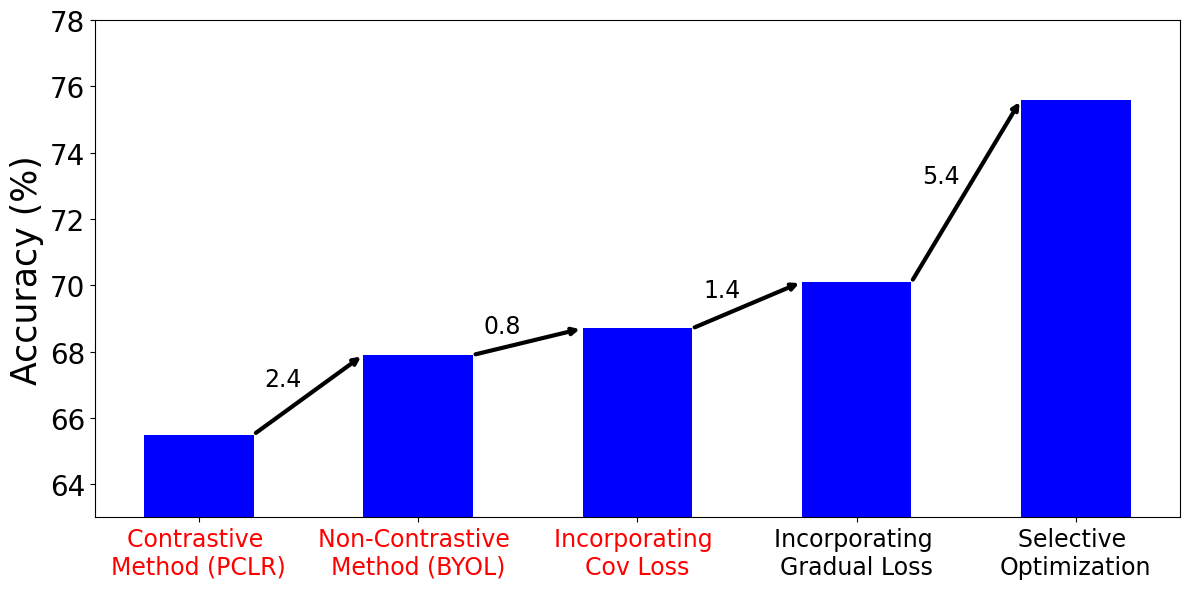}\label{fig:abla1}}}
\hfill
\subfloat[LOO MIT-AFIB]
{\fbox{\includegraphics[width=0.5 \textwidth]{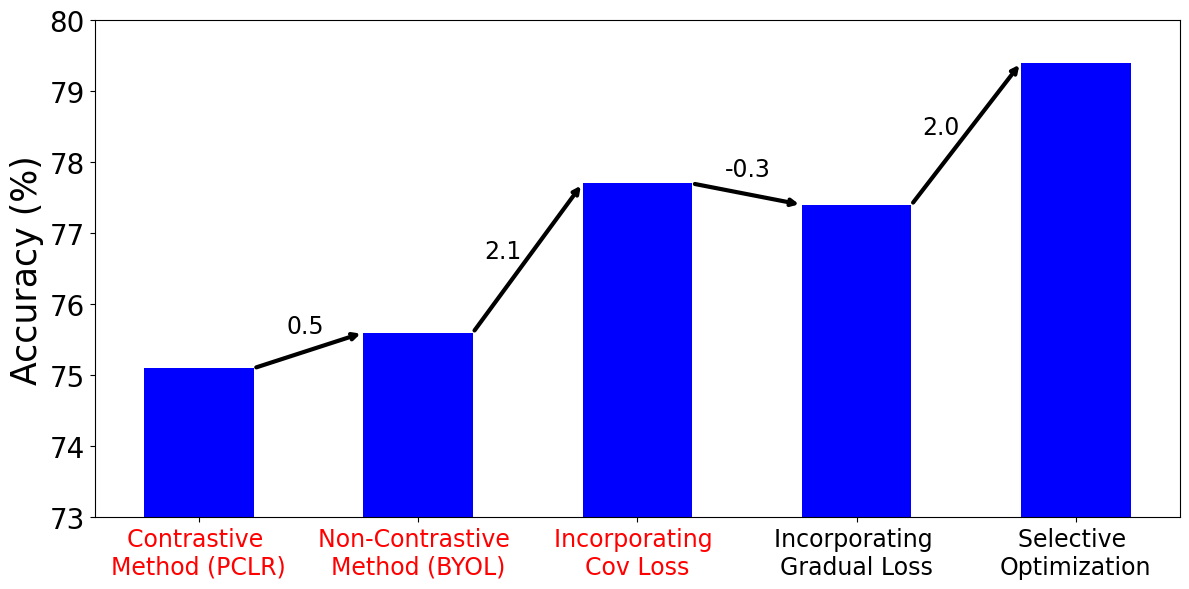}\label{fig:abla2}}}
}
\caption{Effect of incorporating different elements to the training procedure. Components inherited from other existing methods are labeled in red. The impact of adding each component is displayed by the difference in the performance of each downstream task.}
\label{fig:abla}
\end{figure*}

\subsection{Discussion of the Results} 
This study has demonstrated that by following a non-contrastive methodology in addition to introducing a training objective ($\mathcal{L}_{gra}$) to focus on the dynamic patterns, both static and dynamic natures of time series data are encoded by the model (Figure \ref{fig:pca}). It leads to a performing model, which can be used in a wide spectrum of downstream tasks, using different types of datasets (Tables \ref{tab:afib}, \ref{tab:gender_table} and \ref{tab:physio}).
Compared to related methods, \ac{DEAPS} shows a superior performance.\\

\noindent These results show substantial support for the hypothesis: (i) the use of Contrastive Learning techniques, by introducing an objective that prioritizes that the representations of different recordings are far apart, causes the model to neglect the dynamic patterns, (ii) Dynamic patterns are much less evident than static ones, so avoiding the use of negative pairs is not enough. An additional objective function is needed to lead the model to capture them. (iii) By effectively capturing these dynamic patterns, the model will acquire a better generalization capability by just considering a few labeled records.

\section{Ablation Study}\label{sec:ablation}
This ablation study evaluates the effect of incorporating the different parts of the proposed method. In addition, the effect of the effect of the configuration of the two hyperparameters of the method i.e. window size and number of features that are considered in the Selective Optimization. 

\subsubsection{Role of the Different Components}
Starting with contrastive methods, DEAPS changes this contrastive goal to the student-teacher architecture proposed by \ac{BYOL} work.  In addition, it incorporates the covariance term from methods such as \ac{VIC-REG}. In Figure \ref{fig:abla} it can be seen how these two components, which are not part of the contributions of this study, slightly improve these Contrastive Methods. In addition, it shows how the incorporation of the gradual loss, together with the selective optimization, plays a decisive role in obtaining the results with which \ac{DEAPS} significantly improves existing \ac{SOTA} methods.

\subsubsection{Role of the Different Hyperparameters}
As it is described in Section \ref{sec:gra_loss} The size of the spacing window must be sufficiently broad to capture signal variations but sufficiently limited to ensure that the direction of these variations is constant. Table \ref{tab:win_abla} shows the impact of this window size. Although the proposed configurations improve existing methods, the best performance level is achieved by considering inputs belonging to a two-minute window size.

\begin{table}[H]
\centering
\resizebox{0.8\columnwidth}{!}{%
\begin{tabular}{|c|cc|}
\hline
\textbf{\begin{tabular}[c]{@{}c@{}}Window Size\\ (Seconds)\end{tabular}} & \textbf{\begin{tabular}[c]{@{}c@{}}MIT-ARR $\rightarrow$ MIT-AFIB \\ Accuracy(\%)\end{tabular}} & \textbf{\begin{tabular}[c]{@{}c@{}}MIT-AFIB LOO \\ Accuracy (\%)\end{tabular}} \\ \hline
90                                                                       & 71.3                                                                                            & 77.3                                                                           \\
120 (Proposed)                                                           & \textbf{75.5}                                                                                   & \textbf{79.4}                                                                  \\
150                                                                      & 69.7                                                                                            & 79.1                                                                           \\ \hline
\end{tabular}}
\caption{Ablation Study on the length of the window size for the $\mathcal{L}_{gra}$ computation.}
\label{tab:win_abla}
\end{table}

\noindent The last aspect to take into consideration is the number of features that are taken into account in each iteration, as described in Section \ref{sec_sel_opt}. Empirically we have determined that they should be 32. Table \ref{fig:abla2} shows the results of different configurations, where it can be seen that the proposal obtains the best performance.  It can be seen that all configurations achieve better performance than existing methods.

\begin{table}[H]
\centering
\resizebox{0.8\columnwidth}{!}{%
\begin{tabular}{|c|cc|}
\hline
\textbf{\# Features} & \textbf{\begin{tabular}[c]{@{}c@{}}MIT ARR $\rightarrow$ MIT AFIB \\ Accuracy(\%)\end{tabular}} & \textbf{\begin{tabular}[c]{@{}c@{}}MIT-AFIB LOO \\ Accuracy (\%)\end{tabular}} \\ \hline
16                   & 69.3                                                                                            & 79.2                                                                           \\
32 (Proposed)        & \textbf{75.5}                                                                                   & \textbf{79.4}                                                                  \\
48                   & 72.4                                                                                            & 78.9                                                                           \\ \hline
\end{tabular}}
\caption{Ablation Study on the number of features to be considered during the Selective Optimization}
\label{tab:fea_abla}
\end{table}

\section{Contributions}
This study presents compelling evidence elucidating the tendency of Contrastive Learning objectives to induce neglect of subtle changes within quasiperiodic time series data. Addressing this limitation, we introduce \ac{DEAPS} as a novel \ac{SSL} method, distinct from conventional SOTA Contrastive Learning approaches. By dispensing with negative pairs, our method mitigates the model's susceptibility to capture shifts and integrates the novel $\mathcal{L}_{gra}$ function. This function guides the model to effectively encode dynamic patterns. Consequently, the model's capacity to generalize across different classes is demonstrated. The practical impact of this advancement is highlighted through a notable improvement of up to 10\% compared to existing methods.\\

\noindent We believe that this characterization of physiological signals in static and dynamic patterns as well as the incorporation of a specific training objective for dynamic ones will provide inspiration for future Time Series \ac{SSL} methods.

\subsubsection{Limitations}
%While we assert the potential applicability of \ac{DEAPS} to physiological data in general, we acknowledge that our experiments have been limited to the analysis of \ac{ECG} data. 

Only one database (SHHS) has been used to pre-train the model. Using time strips from different recordings for the computation of $\mathcal{L}_{sim}$ is found to be the best strategy. SHHS is the only database we have found with multiple recordings from the same subject. However, we believe we have mitigated this lack of distinct datasets for pre-training by involving multiple datasets in the evaluation.

\appendix

%% The file named.bst is a bibliography style file for BibTeX 0.99c

\bibliographystyle{named}
\bibliography{ijcai24}

\newpage 
\appendix

\begin{figure}[t]
    \centering
    \fbox{\includegraphics[width=0.9\linewidth]{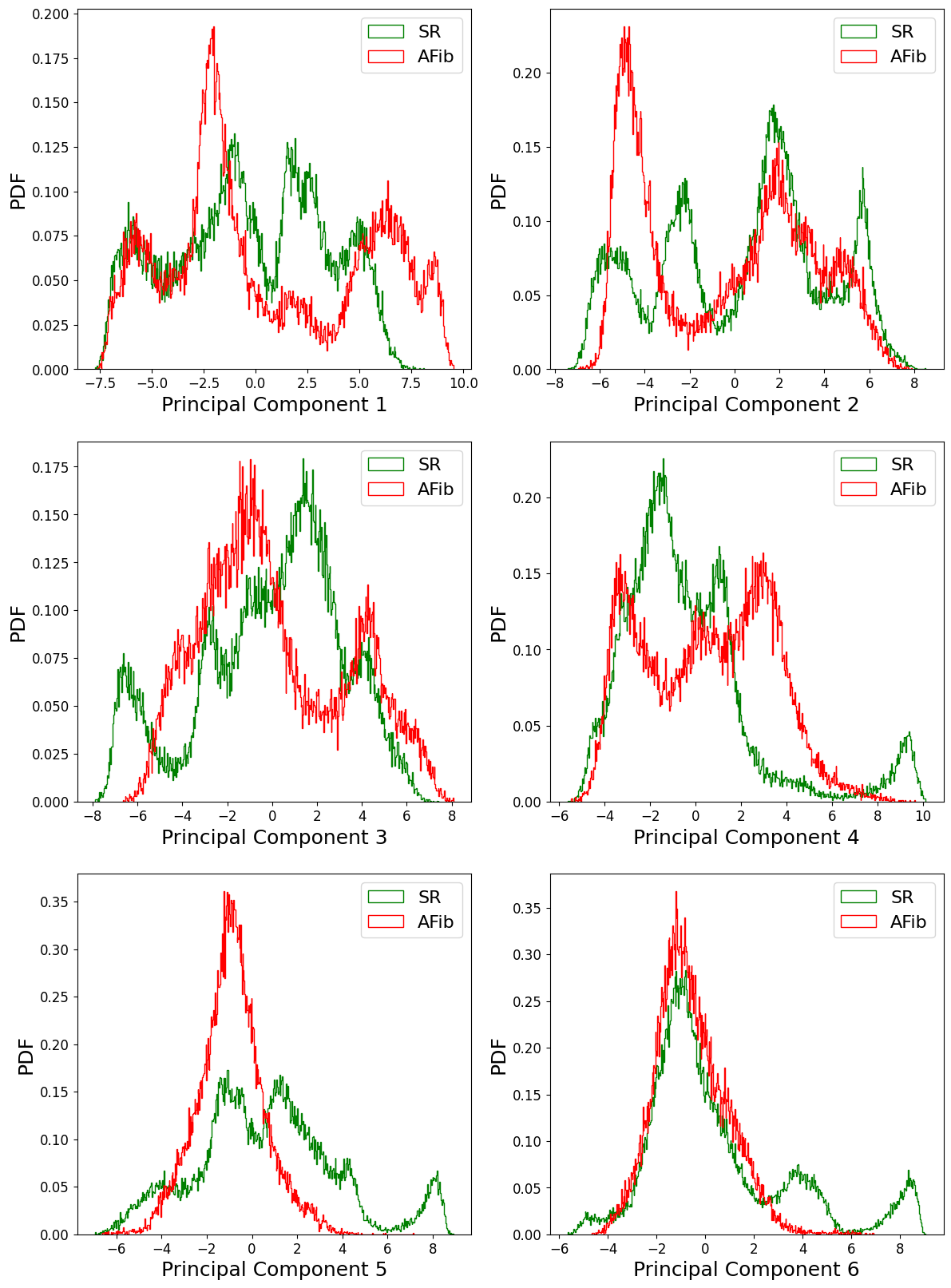}}
    \caption{Episode-based PCA on the representations obtained by the PCLR Contrastive Learning method. None of the first 6 components have significantly different values between SR and AFib episodes.}
    \label{fig:pclr_pca}
\end{figure}

\begin{figure}[t]
\centering
{\fbox{\includegraphics[width=0.9\linewidth]{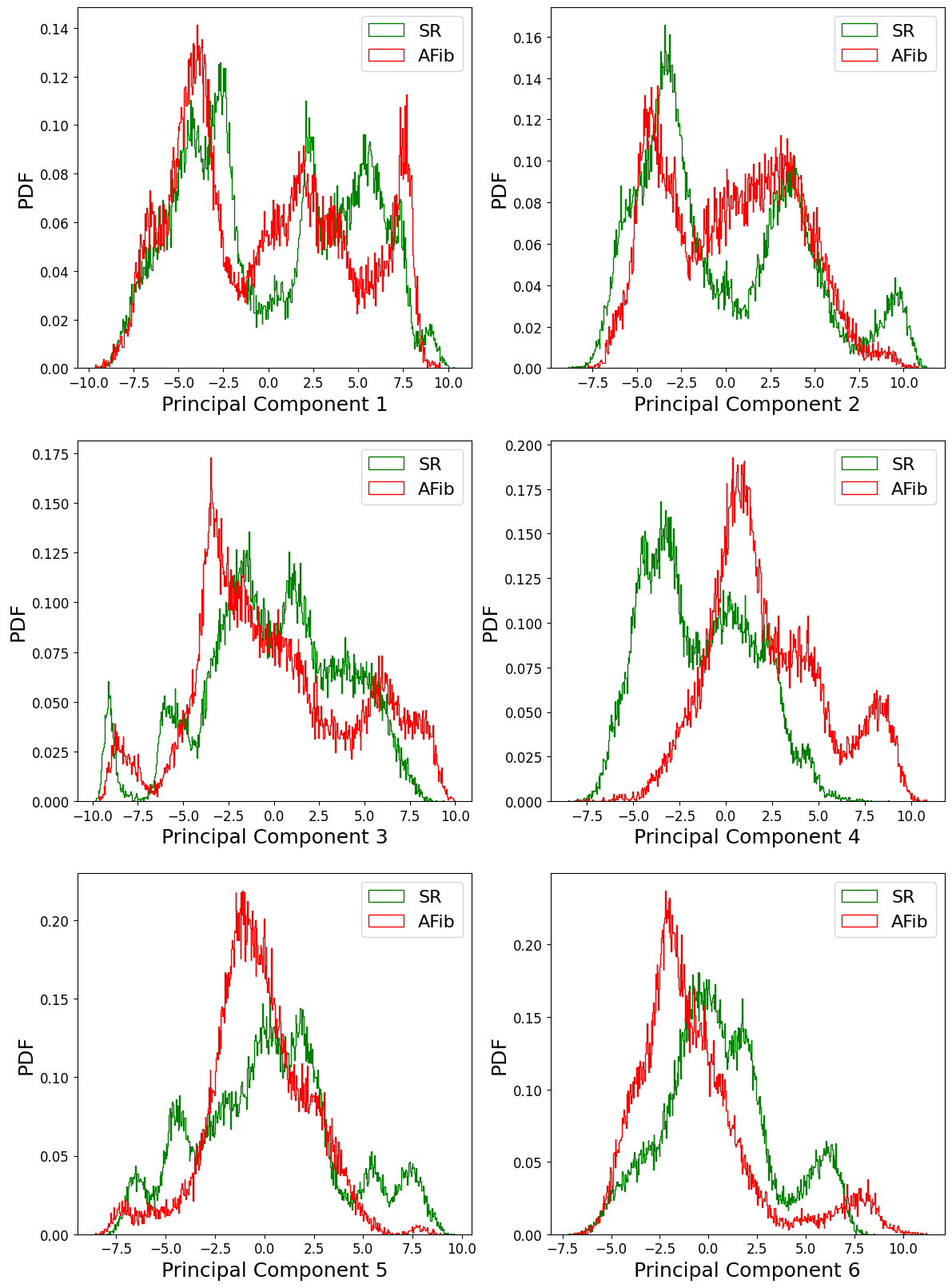}}}
\caption{Episode-based PCA on the representations obtained by  the CLOCS method. None of the first 6 components have significantly different values between SR and AFib episodes.} 
\label{fig:clocs_pca}
\end{figure}

\begin{figure}[H]
\centering
{\fbox{\includegraphics[width=0.9\linewidth]{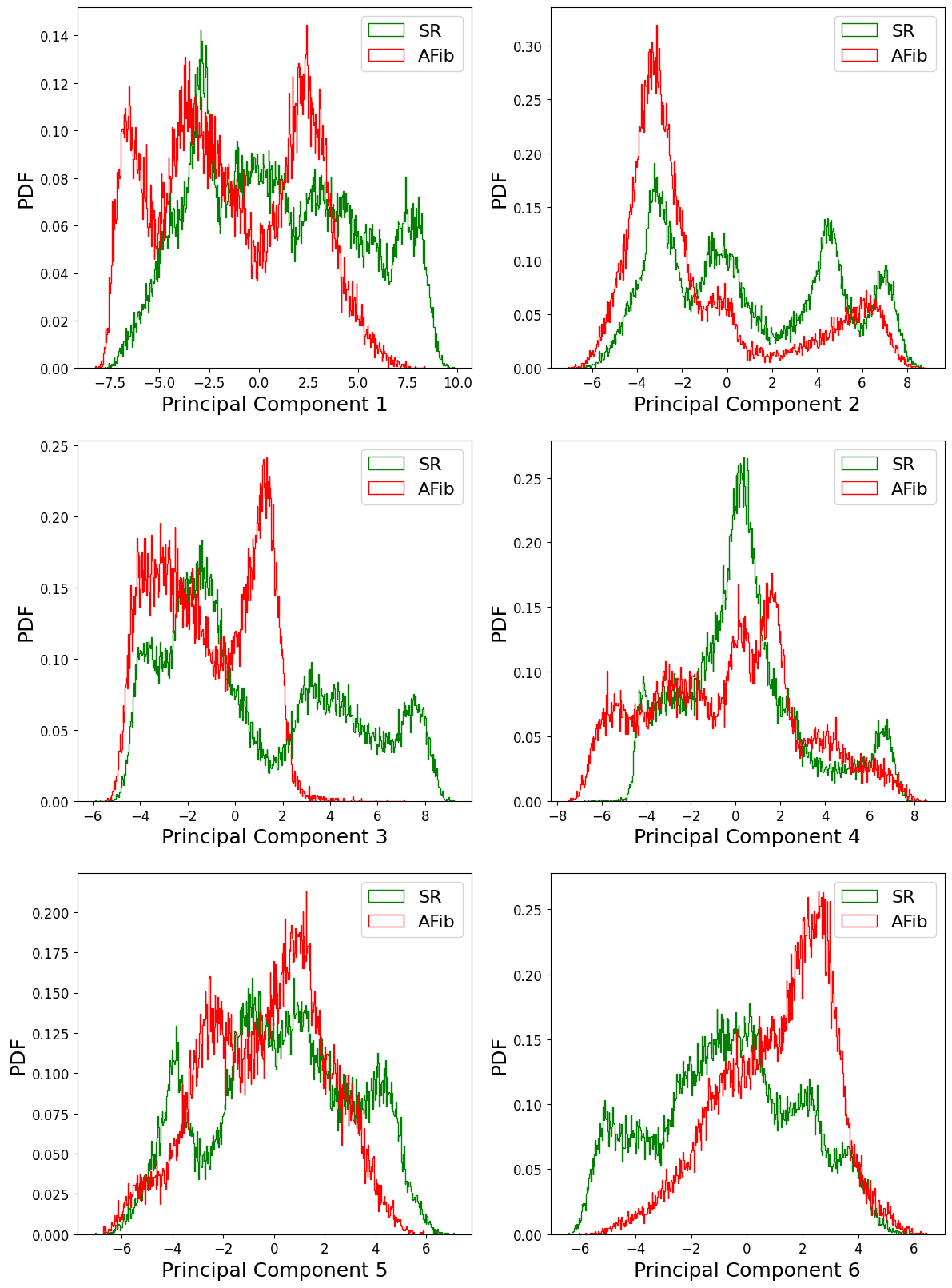}}}
\caption{Episode-based PCA on the representations obtained by baseline non-contrastive method. None of the first 6 components have significantly different values between SR and AFib episodes.} 
\label{fig:byol_pca}
\end{figure}

\section{Contrastive Learning PCA Analysis}

The same exploration of the representations carried out in Section \ref{sec:eval} has been performed with the representations obtained by optimizing the model with the others Contrastive Learning methods. Figures \ref{fig:pclr_pca} and \ref{fig:clocs_pca} show the results of the PDF functions for the values of each of the first 6 Principal Components. It can be seen that, unlike the results obtained by DEAPS, the Contrastive Learning method does not compute significantly different values for each episode.

Additionally, the same experiment was repeated with the BYOL method used as baseline. Figure \ref{fig:byol_pca} shows how simply dispensing with the negative-pairs is not sufficient for a linear combination of features in the representation to show significantly different values for different episodes. This highlights the role of the $\mathcal{L}_{gra}$ in capturing these dynamic patterns.
\section{Data Preprocessing} \label{app:data}
To ensure complete reproducibility of this work, this section presents a detailed description of the preprocessing steps employed for the training and evaluation databases utilized in the proposed method.
\subsection{\ac{SHHS} Data Selection}
Only the subjects which appear in both recording cycles are used during the training procedure. This leads to 2643 subjects. ECG signals are extracted from the \ac{PSG} recordings. The quality of every 10 seconds-data strips has been evaluated with the algorithm proposed by Zhao and Zhang \cite{ecg_quality}.

We use SHHS since it contains two records belonging to the same subject. This makes this specific database special, and this is the reason that it has been the only database used during the optimization.

\subsection{Data cleaning}
In addition, all signals from the utilized datasets were resampled to a frequency of 100Hz. Then, a $5^{th}$ order butterworth high-pass filter with a cutoff frequency of 0.5Hz was applied to eliminate any DC-offset and baseline wander. Finally, each dataset underwent normalization to achieve unit variance, ensuring that the signal samples belong to a $\mathcal{N}(0, 1)$ distribution. This normalization process aimed to mitigate variations in device amplifications that may have occurred during the data collection.

\section{The algorithm}
In order to guarantee the reproducibility of the presented method, the DEAPS pseudocode has been detailed. Algorithm \ref{algo} details, step by step, each of the parts of the method to facilitate its implementation.

\begin{algorithm*}[t]
\SetAlgoLined
\KwIn{
    \\  \hspace{0.5cm} \textit{D}, \textit{K} and \textit{N} \Comment{Set of time series, Number of iterations and Batch Size}
    \\ \hspace{0.5cm} $\mathcal{F}(x)$ and $\mathcal{F\prime}(x)$ \Comment{Student Encoder and Teacher Encoder}
    \\ \hspace{0.5cm}  $\mathcal{G}(h)_{s}$ and $\mathcal{G}(h)_{d}$ \Comment{Student Static and Dynamic Projectors}
    \\ \hspace{0.5cm} $\mathcal{G\prime}(h)_{s}$ and  $\mathcal{G\prime}(h)_{d}$ \Comment{Teacher Static and Dynamic Projectors}
    \\ \hspace{0.5cm}  $\mathcal{Q}(z)_{s}$ and $\mathcal{Q}(z)_{d}$ \Comment{Student Static and Dynamic Predictors}
    \\ \hspace{0.5cm} $\theta$ , and $\xi$ \Comment{Student and Teacher Parameters}
    \\ \hspace{0.5cm} \textit{opt} and $\tau$ \Comment{Optimizer and EMA update parameter}
    %\\ \hspace{0.5cm} $\mathcal{PAR}$ \Comment{Ponderate Average Representation}
    \\ \hspace{0.5cm} $\mathcal{L}_{sim}$ and $\mathcal{L}_{gra}$ \Comment{Similarity and Gradual Loss Functions} 
    \\ \hspace{0.5cm}$\mathcal{M}_{gra}(\mathcal{Q}_d(z_{t - i}), \mathcal{Q}_d(z_{t + j}))$\Comment{Gradual Mask}
    \\ \hspace{0.5cm} $\mathcal{L}_c(z)$  and  $\alpha$ \Comment{Covariance Loss and Covariance Coefficient}
    \\ \hspace{0.5cm} $w_{size}$  \Comment{Windows Size}
    }
\vspace{0.25cm}
\For{$k \gets 0$ \KwTo $K$}{
        $\mathcal{B} \gets {\{X_1, X_2^{t - i}, X_2^{t}, X_2^{t + j} \in \textit{D}\}_{n = 0} ^ N}$ 
        \Comment{Sample $X_1, X_2^{t - i}, X_2^{t}, X_2^{t + j}$ from dataset}\\ 
        \vspace{0.2cm}
        \For{$X_1, X_2^{t - i}, X_2^{t}, X_2^{t + j} \in \mathcal{B}$}{
        \vspace{0.2cm}
        $h_1, h_2^{t - i}, h_2^{t}, h_2^{t + j} \gets \mathcal{F}(X_1,X_2^{t - i}, X_2^{t}, X_2^{t + j})$ 
        \Comment{Student Encoder Representations}\\
        \vspace{0.1cm}
        $\eta_1, \eta_2^{t - i}, \eta_2^{t}, \eta_2^{t + j} \gets \mathcal{F\prime}(X_1, X_2^{t - i}, X_2^{t}, X_2^{t + j})$ \Comment{Teacher Encoder Representations}\\
        \vspace{0.2cm}
        $z_1^s, z_2^{s} \gets \mathcal{G}_{s}(h_1, h_2^{t})$ \Comment{Student Static Projections}\\
        \vspace{0.1cm}
        $z_{t - i}^{d}, z_{t}^{d}, z_{t + j}^{d} \gets \mathcal{G}_{d}(h_2^{t - i}, h_2^{t}, h_2^{t + j})$ \Comment{Student Dynamic Projections}\\
        \vspace{0.2cm}
        $\upzeta_1^s, \upzeta_2^{s} \gets \mathcal{G\prime}_{s}(\eta_1, \eta_2^{t})$ \Comment{Teacher Static Projections}\\
        \vspace{0.1cm}
        $\upzeta_{t - i}^{d}, \upzeta_{t}^{d}, \upzeta_{t + j}^{d} \gets \mathcal{G\prime}_{d}(\eta_2^{t - i}, \eta_2^{t}, \eta_2^{t + j})$ \Comment{Teacher Dynamic Projections}\\
        \vspace{0.6cm}
         $m_{gra} \gets \mathcal{M}_{gra}(\mathcal{Q}_d(z_{t - i}), \mathcal{Q}_d(z_{t + j}))$ \Comment{Gradual Mask from Dynamic Projections}\\

        \vspace{0.5cm}
        $\mathbf{l}_n^{sim} \gets  0.5\cdot (\mathcal{L}_{sim}(\mathcal{Q}_s({z}_{1}^s), \mathbf{\upzeta}_{2}^s) + $  $\mathcal{L}_{sim}(\mathcal{Q}_s({z}_{2}^s), \mathbf{\upzeta}_{1}^s)) $\Comment{Static Loss}\\
             
        \vspace{0.2cm}
        
        $\mathbf{l}_n^{gra} \gets 0.5 \cdot(\mathcal{L}_{gra}(m_{gra}\cdot\mathcal{Q}_d({z}_{t - i}^d), m_{gra}\cdot{\upzeta}_{t}^d, m_{gra}\cdot\mathcal{Q}_d({z}_{t + j}^d)) + $ 
        \\ \hspace{1.9cm} $\mathcal{L}_{gra}(m_{gra}\cdot{\upzeta}_{t - i}^d, m_{gra}\cdot\mathcal{Q}_d({z}_{t}^d), m_{gra}\cdot\mathbf{\upzeta}_{t + j}^d)) $\Comment{Gradual Loss}\\

        \vspace{0.2cm}
        $\mathbf{l}_n^{Cov} \gets \alpha \cdot \mathcal{L}_c(z^s, z^d)$\Comment{Covariance Term Loss}\\
        \vspace{0.2cm}
    }
    \vspace{0.2cm}
    $\partial \theta \gets \sum_{n=0}^{N} (\partial_{\theta} \mathbf{l}_n^{Static} + \partial_{\theta} \mathbf{l}_n^{STD} + \partial_{\theta} \mathbf{l}_n^{Cov})$ \Comment{Compute loss gradients for $\theta$} \\

    \vspace{0.2cm}
    $\theta \gets \textit{opt}(\theta, \partial_{\theta})$ \Comment{Update Student Parameters}\\
    $\xi \leftarrow \tau \cdot \xi+(1-\tau) \cdot \theta$ \Comment{Update Teacher Parameters}\\
    }
\caption{\acf{DEAPS}}
\label{algo}
\end{algorithm*}

\end{document}